\def\year{2022}\relax
\documentclass[letterpaper]{article} 
\usepackage{aaai22}  
\usepackage{times}  
\usepackage{helvet}  
\usepackage{courier}  
\usepackage[hyphens]{url}  
\usepackage{graphicx} 
\usepackage{natbib}  
\usepackage{caption} 
\DeclareCaptionStyle{ruled}{labelfont=normalfont,labelsep=colon,strut=off} 
\usepackage{algorithm}
\usepackage{algorithmicx}
\usepackage[noend]{algpseudocode}
\usepackage{amsmath}
\usepackage{amssymb}
\usepackage{amsthm}
\usepackage{bbm}
\usepackage{bm}
\usepackage{dirtytalk}
\usepackage{dsfont}
\usepackage{enumerate}
\usepackage{graphicx}
\usepackage{mathtools}
\usepackage{subfigure}
\usepackage{times}
\usepackage{xspace}

\usepackage[capitalize,noabbrev]{cleveref}

\usepackage[disable]{todonotes}

\newcommand{\commentout}[1]{}
\newcommand{\junk}[1]{}

\newcommand{\cA}{\mathcal{A}}

\newcommand{\cD}{\mathcal{D}}

\newcommand{\cL}{\mathcal{L}}
\newcommand{\cN}{\mathcal{N}}
\newcommand{\cP}{\mathcal{P}}
\newcommand{\cV}{\mathcal{V}}
\newcommand{\cX}{\mathcal{X}}

\newcommand{\realset}{\mathbb{R}}

\newcommand{\Esub}[2]{\mathbb{E}_{#1} \left[#2\right]}

\newcommand{\abs}[1]{\left|#1\right|}

\newcommand{\I}[1]{\mathds{1} \! \left\{#1\right\}}

\newcommand{\normw}[2]{\|#1\|_{#2}}

\newcommand{\set}[1]{\left\{#1\right\}}

\newcommand{\T}{^\top}
\newcommand{\trace}[1]{\mathrm{tr} \left(#1\right)}

\DeclareMathOperator*{\argmin}{arg\,min\,}

\DeclareMathOperator{\rank}{rank}

\mathchardef\mhyphen="2D

\newcommand{\aopt}{\ensuremath{\tt A\mhyphen optimal}\xspace}
\newcommand{\eopt}{\ensuremath{\tt E\mhyphen optimal}\xspace}
\newcommand{\uniform}{\ensuremath{\tt Prod}\xspace}
\newcommand{\markbaseline}{\ensuremath{\tt QR}\xspace}

\title{Optimal Probing with Statistical Guarantees for Network Monitoring at Scale}
\author{
  Muhammad Jehangir Amjad, \textsuperscript{\rm 1}
  Christophe Diot, \textsuperscript{\rm 1}
  Dimitris Konomis, \textsuperscript{\rm 2} \\
  Branislav Kveton, \textsuperscript{\rm 3}\footnote{This work was done prior to joining Amazon.}
  Augustin Soule, \textsuperscript{\rm 1}
  Xiaolong Yang \textsuperscript{\rm 1}
}
\affiliations{
  \textsuperscript{\rm 1} Google,
  \textsuperscript{\rm 2} MIT,
  \textsuperscript{\rm 3} Amazon
}

\begin{document}

\maketitle

\begin{abstract}
Cloud networks are difficult to monitor because they grow rapidly and the budgets for monitoring them are limited. We propose a framework for estimating network metrics, such as latency and packet loss, with guarantees on estimation errors for a fixed monitoring budget. Our proposed algorithms produce a distribution of probes across network paths, which we then monitor; and are based on A- and E-optimal experimental designs in statistics. Unfortunately, these designs are too computationally costly to use at production scale. We propose their scalable and near-optimal approximations based on the Frank-Wolfe algorithm. We validate our approaches in simulation on real network topologies, and also using a production probing system in a real cloud network. We show major gains in reducing the probing budget compared to both production and academic baselines, while maintaining low estimation errors, even with very low probing budgets.
\end{abstract}

\section{Introduction}
\label{sec:introduction}

Network monitoring is critical to many management and traffic engineering tasks, which enable high service availability. The research community has enthusiastically on-boarded the challenge of designing scalable network monitoring systems \cite{10.1145/2934872.2934906, 10.5555/2930611.2930632, 10.1145/3230543.3230544, 10.1145/3341302.3342076, 10.1145/2829988.2787483, 10.1145/2934872.2934879, 10.1145/3230543.3230559}, an effort which generally relies on the ability to (i) collect exhaustive data in a timely manner, (ii) optimally store the data or its summary, and (iii) perform efficient computations to accurately estimate network metrics. Despite the appealing properties, these systems are hard to deploy in production for a number of reasons. First, the geographical and infrastructural scale of cloud networks is prohibitively large and usually growing at a rapid pace. Secondly, routers and hosts in the network often have limited memory and CPU resources. Within hosts, the goal is to preserve as many resources as possible for revenue generation. As a result, there is no acceptable amount of CPU or memory that can be reserved for monitoring, which translates to \say{using as little as possible}.

In this work, instead measuring all network traffic, we show how to judiciously collect best possible measurements at a fixed budget. In a sense, we build better estimators of network metrics from \emph{better allocated measurements}, not necessarily more. We approach this problem as A- and E-optimal experimental designs in statistics \cite{pazman86foundations,atkinson92optimum,pukelsheim93optimal}. The designs generate a probing distribution over network paths, which we later probe. After we collect the measurements of the probed paths, we use them to estimate a model of network metrics with statistical guarantees.

This work has its roots in network tomography \cite{10.1145/511399.511338}, the ability to infer network metrics from probes sent in the network. It is also related to optimal monitor placement \cite{4068080}, where to monitor and at what rate, to accurately measure metrics with limited impact on network resources. The closest related work is \citet{DBLP:conf/sigmetrics/ChuaKC05}, who propose a method to select approximately the best subset of network paths to measure. We compare to this baseline in our experiments. Related work is reviewed in more detail in \cref{sec:related work}.

We make the following contributions. First, we propose probing distributions for network monitoring that provably minimize estimation errors of network metrics at a fixed monitoring budget (\cref{sec:optimal designs}). The distributions are optimal but computationally costly to obtain. Second, we propose near-optimal approximations to computing the distributions that are orders of magnitude faster to compute (\cref{sec:practical near-optimal designs}). This reduces the computational cost of our solutions from many hours to just a few minutes, and allows deployment in near real-time in practice. Finally, we evaluate our approach by simulations (\cref{sec:synthetic experiments}) and probing in a real cloud production network (\cref{sec:real-world experiments}). We produce more accurate metric estimates than all baselines for similar probing budgets. Therefore, our solutions are not only computed fast, but are also accurate. Most importantly, even with low budgets, our estimation errors are low, which makes our approach applicable to many telemetry tasks.

\section{Framework}
\label{sec:framework}

We cast the problem of network monitoring as estimating a function $f_\ast: \cX \to \realset$, which maps $d$-dimensional \emph{feature vectors} $x \in \cX$ to a \emph{network metric} $y \in \realset$, where $\cX \subseteq \realset^d$ is a set of all $d$-dimensional feature vectors. An example of $f_\ast$ is the \emph{latency of a path} in a network with $d$ edges. In this case, $x \in \cX$ are edge indicators in a path, $y = f_\ast(x)$ is the expected latency of that path, and $\cX \subseteq \set{0, 1}^d$ is the set of all paths.

\subsection{General Approach}
\label{sec:general approach}

The function $f_\ast$ is estimated from a dataset of size $n$, $\cD = \set{(x_i, y_i)}_{i = 1}^n$, where $x_i \in \realset^d$ is the feature vector of the $i$-th example and $y_i \in \realset$ is its noisy observation. In network monitoring, $x_i$ in $\cD$ can be chosen, we can decide what to measure. In this work, we study the problem of collecting a \say{good} dataset $\cD$, such that $\hat{f}$ estimated from $\cD$ approximates $f_\ast$ well. This problem is challenging, since $\hat{f}$ depends on the measurements $y_i$ in $\cD$, which are unknown before $\cD$ is collected. Perhaps surprisingly, this problem can be recast as determining a distribution of feature vectors $x$ in $\cD$ that would lead to a good approximation $\hat{f}$ with guarantees, after $f_*$ is measured at those feature vectors. This problem is known as the \emph{optimal experimental design} \cite{pazman86foundations,atkinson92optimum,pukelsheim93optimal}.

\begin{algorithm}[t]
  \caption{Estimating $\hat{f}$ from probes.}
  \label{alg:probing}
  \begin{algorithmic}[1]
    \State \textbf{Inputs:} Budget of $n$ measurements
    \Statex
    \State Determine probing distribution $\alpha = (\alpha_x)_{x \in \cX}$
    \State $\cD \gets \emptyset$
    \For{$i = 1, \dots, n$}
      \State Set $x_i \gets x$ with probability $\alpha_x$
      \State Probe $x_i$ and observe $y_i$
      \State $\cD \gets \cD + (x_i, y_i)$
    \EndFor
    \State Learn $\hat{f}$ from the collected dataset $\cD$
    \Statex
    \State \textbf{Outputs:} Learned function $\hat{f}$
  \end{algorithmic}
\end{algorithm}

Our approach is outlined in \cref{alg:probing}. In line $2$, we determine the \emph{probing distribution} $\alpha = (\alpha_x)_{x \in \cX}$ of feature vectors, where $\alpha_x$ is the \emph{probing probability} of feature vector $x$. This is the main optimization problem studied in this work. In lines $3$-$7$, we collect a dataset $\cD$ of size $n$, where each $x$ appears $\alpha_x n$ times in expectation. Finally, in line $8$, we learn an approximation $\hat{f}$ to $f_*$ from $\cD$.

\subsection{Optimized Errors}
\label{sec:optimized errors}

We study two notions of how well $\hat{f}$ in \cref{alg:probing} approximates $f_*$. The first is the \emph{worst case}, where $\hat{f}$ approximates $f_\ast$ well for all $x \in \cX$. One metric that captures this notion is the maximum squared error of function $f$,
\begin{align}
  \textstyle
  \cL_{\max}(f)
  = \max_{x \in \cX} (f(x) - f_\ast(x))^2\,.
  \label{eq:maximum error}
\end{align}
In our latency prediction example, $\cX$ would be the set of all paths and $\cL_{\max}(f)$ would be the maximum squared error of estimated path latencies $f$.

We also consider the \emph{average case}. Specifically, let $\cP$ be a distribution over feature vectors in $\cX$. Then we may want to optimize the average error over $x \sim \cP$,
\begin{align}
  \cL_\mathrm{avg}(f)
  = \Esub{x \sim \cP}{(f(x) - f_\ast(x))^2}\,.
  \label{eq:average error}
\end{align}
In our latency prediction example, $\cP$ would be a distribution over all paths $\cX$ and $\cL_{\mathrm{avg}}(f)$ would be the mean squared error of estimated path latencies $f$, weighted by $\cP$.

\section{Optimal Designs}
\label{sec:optimal designs}

To optimize errors \eqref{eq:maximum error} and \eqref{eq:average error}, we focus on a class of functions $f: \cX \to \realset$ where $\abs{f(x) - f_\ast(x)}$ can be bounded by an expression that only depends on the features, and not on the measurements. One such class are linear models.

\subsection{Linear Models}
\label{sec:linear models}

In linear models, the unknown value $f_*(x)$ is linear in $x$ and given by $f_\ast(x) = x\T \theta_\ast$, where $x \in \realset^d$ is a known feature vector and $\theta_\ast \in \realset^d$ is an unknown parameter vector. The unknown $\theta_\ast$ is estimated from noisy observations
\begin{align*}
  y
  = f_\ast(x) + \epsilon
  = x\T \theta_\ast + \epsilon\,,
\end{align*}
which are obtained by adding Gaussian noise $\epsilon \sim \cN(0, \sigma^2)$ to the true value. We assume that all examples in dataset $\cD$ are generated in this way. In particular, $y_i = x_i\T \theta_\ast + \epsilon_i$ for all $i \in [n]$, where $\epsilon_i \sim \cN(0, \sigma^2)$ are drawn independently.

The unknown parameter vector $\theta_*$ is typically estimated using least-squares regression \cite{hastie01statisticallearning}, $\hat{\theta} = G^{-1} \sum_{i = 1}^n y_i x_i$, where $G = \sum_{i = 1}^n x_i x_i\T$ is the \emph{sample covariance matrix}. Let $\hat{f}(x) = x\T \hat{\theta}$. It is well-known \cite{durrett_2010} that
\begin{align}
  (\hat{f}(x) - f_\ast(x))^2
  \leq 2 \sigma^2 \log(1 / \delta) x\T G^{-1} x
  \label{eq:squared error to covariance}
\end{align}
holds with probability at least $1 - \delta$ for any $x$. Therefore, the problem of designing a good approximation $\hat{f}$ to $f_\ast$, with a high-probability bound on $(\hat{f}(x) - f_\ast(x))^2$, reduces to designing a good sample covariance matrix $G$, with an upper bound on $x\T G^{-1} x$. Note that $x\T G^{-1} x$ does not depend on the measurements $y_i$ in $\cD$; it only depends on the features $x_i$. This is why we can derive a desirable probing distribution $\alpha$ in \cref{alg:probing} that depends only on $\cX$.

Although linear models are simple, they are useful for estimating latency in networks. In particular, since the latency of a path is the sum of latencies on its edges, it is a linear function. That is, $f_*(x) = x\T \theta_*$ is the mean latency of path $x$, where $\theta_* \in \realset^d$ is a vector of mean latencies of $d$ edges and $x \in \set{0, 1}^d$ is a vector of edge indicators in the path.

\subsection{E-Optimal Design}
\label{sec:e-optimal design}

This section introduces our first approach to optimizing $G$. To present it, we need some terminology from linear algebra, which we introduce next. For any symmetric matrix $M \in \realset^{d \times d}$, $\lambda_i(M)$ denotes its $i$-th largest eigenvalue. We let $\lambda_{\max}(M) = \lambda_1(M)$ and $\lambda_{\min}(M) = \lambda_d(M)$. The trace of $M$ is the sum of its eigenvalues, $\trace{M} = \sum_{i = 1}^d \lambda_i(M)$. We say that $M$ is \emph{positive semi-definite (PSD)}, and denote it by $M \succeq 0$, if $x\T M x \geq 0$ for all $x \in \realset^d$.

Minimization of the maximum error in \eqref{eq:maximum error} reduces to optimizing $G$ as follows. Suppose, without loss of generality, that $\cX$ in \eqref{eq:maximum error} is a unit sphere, $\cX = \set{x: \normw{x}{2} = 1}$. Then
\begin{align*}
  \cL_{\max}(\hat{f})
  \propto \max_{x \in \cX} x\T G^{-1} x
  = \lambda_{\max}(G^{-1})
  = \lambda_{\min}^{-1}(G)\,.
\end{align*}
The first step is from the definition of $\cL_{\max}$ in \eqref{eq:maximum error} and \eqref{eq:squared error to covariance}, where we omit constant $2 \sigma^2 \log(1 / \delta)$. The first equality is the definition of the maximum eigenvalue, and the second equality is a well-known identity \cite{strang2016introduction}.

Therefore, the minimization of \eqref{eq:maximum error} amounts to maximizing the minimum eigenvalue of $G$. This is known as the \emph{E-optimal design} and can be formulated as a \emph{semi-definite program (SDP)} \cite{vandenberghe99applications}
\begin{align}
  \max & \quad \tau
  \label{eq:e-optimal design} \\
  \textrm{s.t.} & \quad
  \sum_{x \in \cX} \alpha_x x x\T \succeq \tau I_d\,, \quad
  \alpha \in \Delta\,,
  \nonumber
\end{align}
where $\Delta$ is the \emph{set of all distributions over $\cX$}. This SDP has two types of variables. The variable $\tau$ is the minimum eigenvalue of $G$ and it is maximized. The variable $\alpha_x$ is the probing probability of feature vector $x$ (line $2$ in \cref{alg:probing}).

\subsection{A-Optimal Design}
\label{sec:a-optimal design}

Minimization of the average error in \eqref{eq:average error} reduces to optimizing $G$ as follows. Let $\cX$ be the unit sphere (\cref{sec:e-optimal design}) and $\cP$ be a uniform distribution over it. Let $U \Lambda U\T = G^{-1}$ be the eigendecomposition of $G^{-1}$, which is PSD by definition. Then
\begin{align*}
  \cL_\mathrm{avg}(\hat{f})
  & \propto \Esub{x \sim \cP}{x\T G^{-1} x}
  = \Esub{x \sim \cP}{x\T U \Lambda U\T x} \\
  & = \Esub{x \sim \cP}{x\T \Lambda x}
  \propto \sum_{i = 1}^d \lambda_i(G^{-1})
  = \trace{G^{-1}}\,.
\end{align*}
The first step is from the definition of $\cL_\mathrm{avg}$ in \eqref{eq:average error} and \eqref{eq:squared error to covariance}, where we omit constant $2 \sigma^2 \log(1 / \delta)$. The second equality holds since $U$ only rotates $x$ and $\cX$ is a sphere. The next step follows from $\cX$ being a sphere.

Therefore, minimization of \eqref{eq:average error} amounts to minimizing the sum of the eigenvalues of $G^{-1}$, which is equal to its trace. This is known as the \emph{A-optimal design} and can be formulated as the following SDP \cite{vandenberghe99applications}
\begin{align}
  \min & \quad \sum_{i = 1}^d \tau_i
  \label{eq:a-optimal design} \\
  \textrm{s.t.} & \quad
  \forall i \in [d]:
  \begin{bmatrix}
    \displaystyle
    \sum_{x \in \cX} \alpha_x x x\T & e_i \\
    e_i\T & \tau_i
  \end{bmatrix}
  \succeq 0\,, \quad
  \alpha \in \Delta\,,
  \nonumber
\end{align}
where $e_i$ is the $i$-th element of the standard $d$-dimensional Euclidean basis and $\Delta$ is defined as in \eqref{eq:e-optimal design}. This SDP has two types of variables. The variable $\tau_i$ is the $i$-th eigenvalue of $G^{-1}$. Thus $\sum_{i = 1}^d \tau_i = \trace{G^{-1}}$ and it is minimized. The variable $\alpha_x$ is the probing probability of feature vector $x$ (line $2$ in \cref{alg:probing}).

\section{Practical Implementation}
\label{sec:practical near-optimal designs}

The experimental designs in \cref{sec:optimal designs} are sound and well understood. Unfortunately, it is impractical to solve them exactly as SDPs. To address this issue, we propose a general practical solution that combines the strengths of linear programming and gradient descent (\cref{sec:frank-wolfe algorithm}). We apply this solution to both the E-optimal (\cref{sec:frank-wolfe e-optimal design}) and A-optimal (\cref{sec:frank-wolfe a-optimal design}) designs, and evaluate it comprehensively in \cref{sec:synthetic experiments,sec:real-world experiments}. We also show how to apply the optimal designs to generalized linear models in \cref{sec:generalized linear models} and non-linear models in \cref{sec:non-linear models}.

\subsection{Frank-Wolfe Algorithm}
\label{sec:frank-wolfe algorithm}

We solve our SDPs approximately by the Frank-Wolfe algorithm \cite{frank56algorithm}. This results in near-optimal solutions and orders-of-magnitude reduction in run time, as shown in \cref{sec:latency experiments}.

The \emph{Frank-Wolfe algorithm} solves constrained optimization problems of the form
\begin{align}
  \textstyle
  \min_{\alpha \in \cA} f(\alpha)\,,
  \label{eq:constrained minimization}
\end{align}
where $f$ is the optimized function, $\cA \in \realset^d$ is a convex feasible region defined by linear constraints, and $\alpha \in \cA$ is the optimized parameter vector. The key idea is to solve \eqref{eq:constrained minimization} iteratively as a sequence of linear programs (LPs). The iteration $i$ proceeds as follows. The input is a feasible solution $\alpha^{(i)} \in \cA$ from the previous iteration. The objective in \eqref{eq:constrained minimization} is approximated by a linear function at $\alpha^{(i)}$ and the corresponding LP is solved as $\alpha' = g(\alpha^{(i)})$, where
\begin{align}
  \textstyle
  g(\alpha)
  = \argmin_{\tilde{\alpha} \in \cA} \tilde{\alpha}\T \nabla f(\alpha)\,.
  \label{eq:frank-wolfe}
\end{align}
Finally, the algorithm solves a line search problem
\begin{align*}
  \textstyle
  \alpha^{(i + 1)}
  = \argmin_{c \in [0, 1]} f(c \alpha^{(i)} + (1 - c) \alpha')\,,
\end{align*}
where $\alpha^{(i + 1)}$ is the next feasible solution. Roughy speaking, this procedure can be viewed as gradient descent on $f$ where $\alpha^{(i + 1)}$ is guaranteed to be feasible. This is a major advantage over gradient descent \cite{10.5555/993483}, which may require a projection to $\cA$.

The Frank-Wolfe algorithm converges when $f$ and $\cA$ are convex. In our setting, it strikes an elegant balance. On one hand, the LP in \eqref{eq:frank-wolfe} can represent the linear constraints in \eqref{eq:e-optimal design} and \eqref{eq:a-optimal design}. On the other hand, the hard problem of eigenvalue optimization is solved by gradient descent. In what follows, we apply this approach to A- and E-optimal designs.

\subsection{Frank-Wolfe E-Optimal Design}
\label{sec:frank-wolfe e-optimal design}

The E-optimal design (\cref{sec:e-optimal design}) optimizes the minimum eigenvalue of the sample covariance matrix,
\begin{align}
  \textstyle
  \max_{\alpha \in \Delta} \, \lambda_{\min}(G_\alpha)\,,
  \label{eq:e-optimal objective}
\end{align}
where $G_\alpha = \sum_{x \in \cX} \alpha_x x x\T$ is a sample covariance matrix parameterized by probing distribution $\alpha \in \Delta$. Then from the definitions of $\lambda_{\min}$ and $G_\alpha$, we have that
\begin{align*}
  \lambda_{\min}(G_\alpha)
  & = \min_{v \in \realset^d: \normw{v}{2} = 1}
  v\T \left(\sum_{x \in \cX} \alpha_x x x\T\right) v \\
  & = \min_{v \in \realset^d: \normw{v}{2} = 1}
  \sum_{x \in \cX} \normw{v\T x}{2}^2 \, \alpha_x\,.
\end{align*}
Since $\sum_{x \in \cX} \normw{v\T x}{2}^2 \, \alpha_x$ is linear in $\alpha$ for any $v$, $\lambda_{\min}(G_\alpha)$ is concave in $\alpha$. So $- \lambda_{\min}(G_\alpha)$ is convex in $\alpha$ and we can solve \eqref{eq:e-optimal objective} by the Frank-Wolfe algorithm (\cref{sec:frank-wolfe algorithm}) with $\cA = \Delta$ and $f(\alpha) = - \lambda_{\min}(G_\alpha)$. Since $\lambda_{\min}(G_\alpha)$ is linear in $\alpha$,
\begin{align*}
  \nabla f(\alpha)
  = - \nabla \lambda_{\min}(G_\alpha)
  = - \left(\normw{v_{\min}\T x}{2}^2\right)_{x \in \cX}\,,
\end{align*}
where $v_{\min}$ is the eigenvector associated with $\lambda_{\min}(G_\alpha)$.

The time to compute the gradient $\nabla f(\alpha)$ is linear in the number of partial derivatives $\abs{\cX}$. Since this is the most computationally-demanding part of our approximation, its run time is nearly linear in $\abs{\cX}$ and the number of iterations. Therefore, our approach scales to large problems.

\subsection{Frank-Wolfe A-Optimal Design}
\label{sec:frank-wolfe a-optimal design}

The A-optimal design (\cref{sec:a-optimal design}) minimizes the trace of the inverse sample covariance matrix,
\begin{align}
  \textstyle
  \min_{\alpha \in \Delta} \, \trace{G_\alpha^{-1}}\,,
  \label{eq:a-optimal objective}
\end{align}
where both $G_\alpha$ and $\Delta$ are defined as in \cref{sec:frank-wolfe e-optimal design}. Since $\trace{G_\alpha^{-1}}$ is convex in $\alpha$ \cite{10.5555/993483}, we can solve \eqref{eq:a-optimal objective} by the Frank-Wolfe algorithm (\cref{sec:frank-wolfe algorithm}) with $\cA = \Delta$ and $f(\alpha) = \trace{G_\alpha^{-1}}$. Moreover, for any invertible $M \in \realset^{d \times d}$, we have that \cite{IMM2012-03274}
\begin{align*}
  \partial \trace{M^{-1}}
  = \trace{\partial M^{-1}}
  = - \trace{M^{-1} (\partial M) M^{-1}}\,.
\end{align*}
Now we apply this to $M = G_\alpha$ and note that $\partial G_\alpha / \partial \alpha_x = x x\T$. This yields
\begin{align*}
  \nabla f(\alpha)
  = - \left(\trace{G_\alpha^{-1} x x\T G_\alpha^{-1}}\right)_{x \in \cX}\,.
\end{align*}

\section{Synthetic Experiments}
\label{sec:synthetic experiments}

In this section, we evaluate our Frank-Wolfe experimental designs in simulation. The experimental setup is introduced in \cref{sec:general setup} and our latency estimation results are presented in \cref{sec:latency experiments}. We show that our approach achieves better than state-of-the-art results in near real-time, which is the prerequisite for deployment. Additional packet loss experiments are in \cref{sec:packet-loss experiments}. We also study the impact of local probing budgets in \cref{sec:local budget experiments}.

\subsection{General Setup}
\label{sec:general setup}

\noindent \textbf{Network Topologies}. We experiment with three network topologies.\footnote{They are geographical subsets of a global cloud network, similar to the one illustrated at \url{https://i.redd.it/jy6459fajpy21.png}.} A network topology is converted to a network graph comprised of nodes and edges. The nodes in this graph are points of presence (POPs), which are a collection of routing and networking equipment. The edges are the physical links between the POPs. Our network topologies have the following properties:
\begin{itemize}
  \item Topology A: $101$ nodes, $251$ edges, $5\,050$ paths.
  \item Topology B: $82$ nodes, $196$ edges, $3\,321$ paths.
  \item Topology C: $192$ nodes, $439$ edges, $15\,428$ paths.
\end{itemize}
We denote the set of paths in the network by $\cX$ and the number of edges by $d$.

\medskip
\noindent \textbf{Network Metric and Budget}. We experiment with latency, the time for a packet to traverse a path in a network. We assume a global budget of $n$ probes per unit time. Additional packet loss experiments are in \cref{sec:packet-loss experiments}.

\medskip
\noindent \textbf{Evaluation Methodology}. All compared methods operate under a total budget of $n$ measurements. We vary $n$ from $3,000$ to $30,000$. Our choice of $n$ may seem low, since probing budgets may be significantly higher in practice. We choose it to study errors when the budget is scarce and it is important to monitor intelligently.

Each compared method is represented by a probing distribution $\alpha$ (line 2 in \cref{alg:probing}). For each method and budget $n$, we report the average of $300$ independent runs of the following end-to-end experiment:
\begin{enumerate}
  \item Compute a probing distribution $\alpha = (\alpha_x)_{x \in \cX}$ (line 2 in \cref{alg:probing}).
  \item Collect dataset $\cD$ by probing paths $\cX$ according to $\alpha$ (lines 4-7 in \cref{alg:probing}). The noisy observations $y_i$ are generated as described in \cref{sec:latency experiments}.
  \item Estimate model parameters $\hat{\theta} \in \realset^d$ from $\cD$ (line 8 in \cref{alg:probing}) using least-squares regression (\cref{sec:linear models}).
  \item Use $\hat{\theta}$ and $\theta_*$ to compute the errors in \eqref{eq:maximum error} and \eqref{eq:average error}.
\end{enumerate} 

We experiment with the E-optimal design in \cref{sec:frank-wolfe e-optimal design}, which we call \eopt, and the A-optimal design in \cref{sec:frank-wolfe a-optimal design}, which we call \aopt. The number of Frank-Wolfe iterations in both designs is $300$, which is sufficient to obtain near-optimal solutions in all of our experiments.

\medskip
\noindent \textbf{Evaluation Criteria}. We use the \emph{maximum error} in \eqref{eq:maximum error} and \emph{average error} in \eqref{eq:average error} in all experiments. The maximum error is the highest error over a set of paths. The average error is measured on the same paths and reflects the average performance. For the maximum error in \eqref{eq:maximum error}, the set of paths is $\cX$. Therefore, this error corresponds to the most mispredicted path latency. For the average error in \eqref{eq:average error}, $\cP$ is a distribution over paths $\cX$. We define $\cP$ by the following generative process. First, we choose an edge, uniformly at random. Then we choose any path with that edge, uniformly at random. This choice guarantees that any edge appears in a sampled path from $\cP$ with probability at least $1 / d$. As a result, we evaluate the quality of learned models uniformly over all of their parameters.

\medskip
\noindent \textbf{Baselines}. The algorithm of \citet{DBLP:conf/sigmetrics/ChuaKC05}, which we call \markbaseline, is a natural baseline because it also chooses the best paths to probe. It works as follows. First, it computes the singular value decomposition (SVD) of the path-edge matrix $M = U \Sigma V\T$. Next it chooses the first $k = \rank(M)$ columns of $U$, denoted by $U_k$, and applies the pivoting QR decomposition to it. The first $k$ chosen rows of $U_k$ represent the \say{important} rows of $M$, the paths that explain the most information in $M$. The budget is allocated evenly to these paths, $\alpha_x = 1 / k$ for any chosen path $x$.

Our second baseline is \uniform, which allocates the budget evenly to all paths, $\alpha_x = 1 / \abs{\cX}$ for any path $x$. It is a natural point of comparison because this simple baseline is common in practice. The shortcoming of \uniform is that it over-probes links that appear in many paths and under-probes those that appear in a few.

\medskip
\noindent \textbf{Expected Trade-off}. Our approach trades off measurement accuracy and probing budget, since it is not always feasible in large scale networks to measure with the maximum accuracy. We do that with theoretical guarantees. Therefore, when compared to the baselines, we look for \emph{achieving a fixed level of accuracy with lower budget or achieving a higher accuracy with a fixed budget}. Naturally, the higher the probing budget, the higher the accuracy, for both our approach and the baselines.

\medskip
\noindent \textbf{Implementation Details}. All experiments are conducted in Python 3.7. To speed up linear algebra, we run them in TensorFlow \cite{tensorflow}, on a cloud instance with $112$ cores and $392$ GB RAM. This is a standard cloud instance that can be used in production. Therefore, the reported run times in \cref{sec:latency experiments,sec:real-world results} reflect those expected in production.

\subsection{Latency Experiments}
\label{sec:latency experiments}

\begin{figure}[t]
  \centering
  \includegraphics[width=3in]{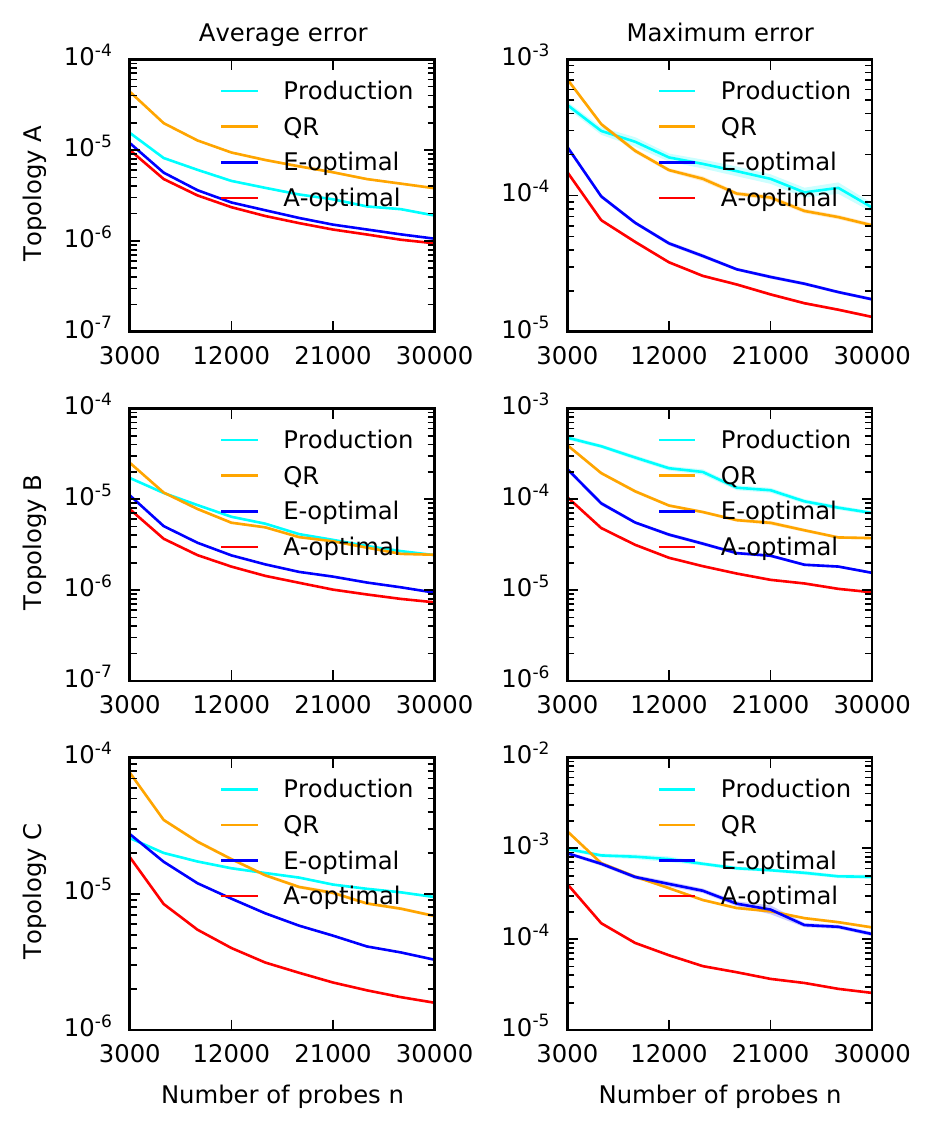}
  \caption{Latency estimation errors in three network topologies. The average errors are in the first column and the maximum errors are in the second.}
  \label{fig:abc latency}
\end{figure}

\noindent \textbf{Latency Simulator}. We denote by $\ell_*(e)$ the mean latency of edge $e$ in seconds. It is defined as the physical distance between the nodes of edge $e$ divided by the speed of light in fiber, which is approximately one third of the speed of light in vacuum. The edge latency in our networks varies from $0.001$s to $0.025$s (topology A), $0.036$s (topology B), $0.011$s (topology C).

Our model (\cref{sec:linear models}) is defined as follows. The mean latency of edge $e$ is $\theta_*(e) = \ell(e)$ and $\theta_* \in \realset^d$ is the vector of mean latencies of all edges. The mean latency of path $x$ is $f_*(x) = x\T \theta_*$, where $x \in \set{0, 1}^d$ is a vector of edge indicators. The \emph{observed latency} of path $x$ is $y = x\T \theta_* + \varepsilon$, where $\varepsilon \sim \cN(0, \sigma^2)$ is independent Gaussian noise. The noise $\varepsilon$ is a \emph{measurement noise} and we set $\sigma = 0.01$, based on prior experience with latency variance.

\noindent \textbf{Results}. All latency estimation errors are reported in \cref{fig:abc latency}. We observe four major trends. First, for all methods, the errors decrease as budget $n$ increases. This is expected, since the quality of latency estimates improves with more probes. Second, the rate of the error decrease, from $n = 3\,000$ to $n = 30\,000$, is about $10$ fold. This follows from how the squared error relates to the sample covariance matrix $G$ in \eqref{eq:squared error to covariance}. In particular, for any probing distribution, $10$ times more probes yield $10$ times higher eigenvalues of $G$, and thus $10$ times lower eigenvalues of $G^{-1}$; which then provide a $10$ times lower error bound in \eqref{eq:squared error to covariance}. Third, the reported errors are acceptable, even for small probing budgets. For instance, in all network topologies, the average squared error of \aopt with budget $n = 30\,000$ is about $10^{-6}$. Thus the absolute path latency is mis-predicted by $10^{-3}$ on average, which is at least an order of magnitude less than the range of the latencies in our networks.

The last trend is that \aopt consistently dominates \eopt, which dominates the baselines \markbaseline and \uniform. That is, \eopt has a lower error than \markbaseline and \uniform for all budgets and attains any fixed error with a lower budget than either of them. We observe this trend for both the average and maximum errors, and over all network topologies. \aopt dominates \eopt in a similar fashion. Perhaps surprisingly, \eopt performs worse than \aopt in minimizing the maximum error. This is because the path with the maximum potential error, the maximum eigenvector of $G^{-1}$, is not in $\cX$.

Finally, we report the run times of our approaches in \cref{tab:run times}. For the E-optimal design, they are about two orders of magnitude lower than if we solved the problems exactly (\cref{sec:e-optimal design}) using CVXOPT \cite{cvxpy}, a popular software for disciplined convex programming. For the A-optimal design, CVXOPT ran out of memory in all problems, because they contain $d$ SDP constraints. In absolute terms, our run times are on the order of a few minutes, and thus suitable for near real-time deployment.

\begin{table}[t]
  \centering
  \begin{tabular}{|l|r|r|r|} \hline
    \textbf{Algorithm / Topology} & A & B & C \\ \hline
    CVXOPT \eopt & 4\,400 & 1\,800 & 31\,700 \\
    CVXOPT \aopt & x & x & x \\
    Frank-Wolfe \eopt & 66 & 57 & 53 \\
    Frank-Wolfe \aopt & 140 & 121 & 219 \\ \hline
  \end{tabular}
  \caption{Run times for computing the probing distributions for topologies A, B, and C.}
  \label{tab:run times}
\end{table}

\section{Real-World Experiments}
\label{sec:real-world experiments}

In \cref{sec:synthetic experiments}, the ground truth and its observations were generated from \say{well-behaved} models. For instance, the mean latency of a path in \cref{sec:latency experiments} is a linear function of its edge latencies, and the latency of that path is observed with Gaussian noise. Real latencies are not always \emph{well-behaved}. Thus we conduct a set of \emph{real-world experiments} with actual observed latencies. Our results show that the Frank-Wolfe solutions can be used in practice. Additional packet loss experiments are in \cref{sec:real-world packet-loss experiments}. 

\noindent \textbf{Network Topology}. All experiments are conducted on a single network topology, which is a random subset of a production network that belongs to a global cloud provider. We cannot reveal the percentage of the network that our topology represents. However, our network is sufficient in size and realistic in complexity: $306$ nodes, $762$ edges, and $39\,316$ paths. This network is larger than any of the networks in \cref{sec:synthetic experiments} and brings additional evidence that our approach can be applied at scale in a cloud network.

\noindent \textbf{Evaluation Methodology}. Our setup mimics a real-world scenario where the probing distribution is recomputed once per unit time, such as one or five minutes. Then the probes are sent and they collect \emph{observations} for each path. Finally, these observations are used to estimate the latency or packet loss of each path in the network, whether or not that path is probed. As in the synthetic experiments, we compare our optimal designs to the \markbaseline and \uniform baselines.

The main challenge in evaluating a real world setting is that the ground truth, the actual latency of a path, is never perfectly known. We determine the ground truth as follows. First, we fix a short time period. We experiment with one and five minutes, and refer to the $1$-minute experiment as \emph{topology M1} and to the $5$-minute experiment as \emph{topology M5}. Next, over that fixed time period, we send $50$ to $100$ probes along all paths $\cX$. In total, this translates to over $4$ million probes. Then we determine the mean latency of any path $x$ as the empirical average of its observations generated by the probes. These correspond to the ground truth latency $f_*(x)$ in \cref{sec:latency experiments}. Our choice of short time periods allows us to determine the ground truth accurately.

\noindent \textbf{Latency Emulator}. We emulate the behavior of our methods on the observations generated by the real probes. As an example, suppose that we have $100$ probes for paths $x_1$, $x_2$, and $x_3$; and the emulated \aopt allocates $6$, $9$, and $0$ observations to these paths. Then the observations are obtained by sampling $6$ random observations from the probes for path $x_1$ and $9$ from those for path $x_2$.

\begin{figure}[t]
  \centering
  \includegraphics[width=3in]{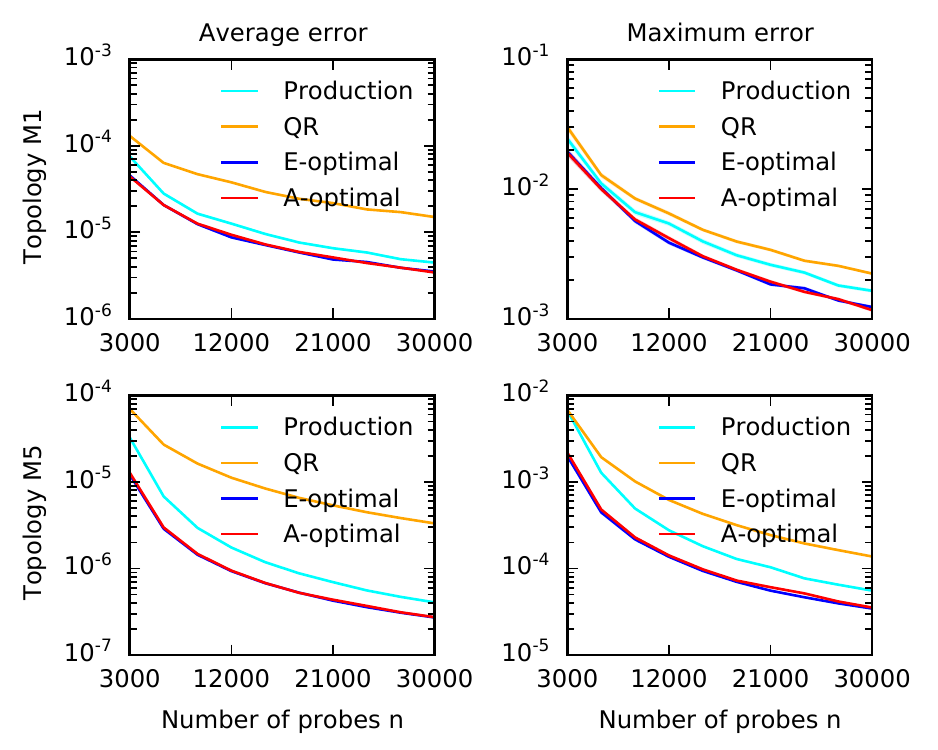}
  \caption{Latency estimation errors in large-scale networks M1 and M5.}
  \label{fig:m15 latency}
\end{figure}

\subsection{Results} 
\label{sec:real-world results}

Our results are reported in \cref{fig:m15 latency}. These results are consistent with the earlier trends in the synthetic experiments (\cref{sec:latency experiments}). Specifically, the errors decrease as budget $n$ increases, and \eopt and \aopt consistently dominate both \markbaseline and \uniform. In all network topologies, the average squared error of A-optimal with budget $n = 30\,000$ is at most $4 \times 10^{-6}$. Thus the absolute path latency is mispredicted by at most $2 \times 10^{-3}$ on average. This is an order of magnitude lower than the estimated quantities, which means that our predictions are reasonably accurate.

The error differences of \eopt, \aopt, \markbaseline, and \uniform are less pronounced than in \cref{sec:synthetic experiments}. One reason is that the ground truth is not idealized. For instance, the path latency is not necessarily linear in its features, which contributes to an additional modeling error. This error can only be eliminated by a better model than linear.

The costs of computing \eopt and \aopt are $346$ and $961$ seconds, respectively. This is six times higher than in the synthetic experiments, for a topology with six times more paths. This is expected, as the run time of our Frank-Wolfe solutions is linear in the number of paths. While this may seem high at first, it is practical. Specifically, since our network routes are usually stable over $10$ minute intervals, we do not expect to recompute the probing distributions more frequently than that.

\section{Conclusions}
\label{sec:conclusions}

We propose a statistical approach to monitoring large-scale network infrastructure, which has provable guarantees on the estimation error for a limited probing budget. We show through simulation and \emph{real network experiments} that our approach yields low errors, even when the probing budget is low. Our approach can be used in any production network to measure any network performance metric, as long as the metric estimation problem can be expressed as a regression problem. Although the main paper focuses solely on linear regression (\cref{sec:linear models}), we discuss GLM and non-linear regression in \cref{sec:generalized linear models,sec:non-linear models}, respectively. Our experimental results show that the cost of computing our probing distributions is not an obstacle to operational deployment. This paper is a good example of taking a statistically sound design and scaling it up to a novel and challenging real-world problem.

Although our initial results are promising, we are only at the initial stages of deploying our system in production networks. One reason is that our current empirical results are limited to only two metrics, latency (\cref{sec:latency experiments,sec:real-world results}) and packet loss (\cref{sec:packet-loss experiments,sec:real-world packet-loss experiments}). In the future, we will evaluate our framework on other telemetry tasks, such as estimating available bandwidth, flow size distribution, or top flows. Such extensions are trivial as long as the metric estimation problem can be formulated as a regression problem. The second reason is that our network topologies are only subsets of the original cloud topology. For now, we plan to use \say{sharding} techniques to divide the full topology into geographical subsets, and then estimate a probing distribution separately for each subset. Finally, since the most computationally demanding part of our approach is linear in the number of paths $\abs{\cX}$ (\cref{sec:frank-wolfe e-optimal design}), such approximations may not be needed at the end to scale up to a Google scale of millions of paths.

\bibliography{References}

\begin{thebibliography}{33}
\providecommand{\natexlab}[1]{#1}

\bibitem[{Atkinson and Donev(1992)}]{atkinson92optimum}
Atkinson, A.; and Donev, A. 1992.
\newblock \emph{Optimum Experimental Designs}.
\newblock Oxford University Press.

\bibitem[{Boyd and Vandenberghe(2004)}]{10.5555/993483}
Boyd, S.; and Vandenberghe, L. 2004.
\newblock \emph{Convex Optimization}.
\newblock USA: Cambridge University Press.

\bibitem[{Bu et~al.(2002)Bu, Duffield, Presti, and
  Towsley}]{10.1145/511399.511338}
Bu, T.; Duffield, N.; Presti, F.~L.; and Towsley, D. 2002.
\newblock Network Tomography on General Topologies.
\newblock \emph{SIGMETRICS Perform. Eval. Rev.}, 30(1): 21--30.

\bibitem[{{Cantieni} et~al.(2006){Cantieni}, {Iannaccone}, {Barakat}, {Diot},
  and {Thiran}}]{4068080}
{Cantieni}, G.~R.; {Iannaccone}, G.; {Barakat}, C.; {Diot}, C.; and {Thiran},
  P. 2006.
\newblock Reformulating the Monitor Placement Problem: Optimal Network-Wide
  Sampling.
\newblock In \emph{2006 40th Annual Conference on Information Sciences and
  Systems}, 1725--1731.

\bibitem[{Chen, Hu, and Ying(1999)}]{chen99strong}
Chen, K.; Hu, I.; and Ying, Z. 1999.
\newblock Strong Consistency of Maximum Quasi-Likelihood Estimators in
  Generalized Linear Models with Fixed and Adaptive Designs.
\newblock \emph{The Annals of Statistics}, 27(4): 1155--1163.

\bibitem[{Chua, Kolaczyk, and Crovella(2005)}]{DBLP:conf/sigmetrics/ChuaKC05}
Chua, D.~B.; Kolaczyk, E.~D.; and Crovella, M. 2005.
\newblock A Statistical Framework for Efficient Monitoring of End-to-End
  Network Properties.
\newblock In \emph{Proceedings of the International Conference on Measurements
  and Modeling of Computer Systems}, 390--391. {ACM}.

\bibitem[{Diamond and Boyd(2016)}]{cvxpy}
Diamond, S.; and Boyd, S. 2016.
\newblock {CVXPY}: A {P}ython-Embedded Modeling Language for Convex
  Optimization.
\newblock \emph{Journal of Machine Learning Research}.
\newblock To appear.

\bibitem[{{Duffield} et~al.(2001){Duffield}, {Lo Presti}, {Paxson}, and
  {Towsley}}]{916283}
{Duffield}, N.~G.; {Lo Presti}, F.; {Paxson}, V.; and {Towsley}, D. 2001.
\newblock Inferring link loss using striped unicast probes.
\newblock In \emph{Proceedings IEEE INFOCOM 2001. Conference on Computer
  Communications. Twentieth Annual Joint Conference of the IEEE Computer and
  Communications Society (Cat. No.01CH37213)}, volume~2, 915--923 vol.2.

\bibitem[{Durrett(2010)}]{durrett_2010}
Durrett, R. 2010.
\newblock \emph{Probability: Theory and Examples}.
\newblock Cambridge Series in Statistical and Probabilistic Mathematics.
  Cambridge University Press, 4 edition.

\bibitem[{Frank and Wolfe(1956)}]{frank56algorithm}
Frank, M.; and Wolfe, P. 1956.
\newblock An Algorithm for Quadratic Programming.
\newblock \emph{Naval Research Logistics Quarterly}, 3(1-2): 95--110.

\bibitem[{Ghita, Argyraki, and Thiran(2013)}]{10.1145/2433140.2433146}
Ghita, D.; Argyraki, K.; and Thiran, P. 2013.
\newblock Toward Accurate and Practical Network Tomography.
\newblock \emph{SIGOPS Oper. Syst. Rev.}, 47(1): 22--26.

\bibitem[{{Ghita} et~al.(2010){Ghita}, {Nguyen}, {Kurant}, {Argyraki}, and
  {Thiran}}]{5461918}
{Ghita}, D.; {Nguyen}, H.; {Kurant}, M.; {Argyraki}, K.; and {Thiran}, P. 2010.
\newblock Netscope: Practical Network Loss Tomography.
\newblock In \emph{2010 Proceedings IEEE INFOCOM}, 1--9.

\bibitem[{Hastie, Tibshirani, and Friedman(2001)}]{hastie01statisticallearning}
Hastie, T.; Tibshirani, R.; and Friedman, J. 2001.
\newblock \emph{The Elements of Statistical Learning}.
\newblock Springer Series in Statistics. New York, NY, USA: Springer New York
  Inc.

\bibitem[{He et~al.(2015)He, Liu, Swami, Towsley, Salonidis, Bejan, and
  Yu}]{10.1145/2796314.2745862}
He, T.; Liu, C.; Swami, A.; Towsley, D.; Salonidis, T.; Bejan, A.~I.; and Yu,
  P. 2015.
\newblock Fisher Information-Based Experiment Design for Network Tomography.
\newblock \emph{SIGMETRICS Perform. Eval. Rev.}, 43(1): 389--402.

\bibitem[{Huang, Lee, and Bao(2018)}]{10.1145/3230543.3230559}
Huang, Q.; Lee, P. P.~C.; and Bao, Y. 2018.
\newblock Sketchlearn: Relieving User Burdens in Approximate Measurement with
  Automated Statistical Inference.
\newblock In \emph{Proceedings of the 2018 Conference of the ACM Special
  Interest Group on Data Communication}, SIGCOMM '18, 576--590.

\bibitem[{Huang, Feamster, and Teixeira(2008)}]{10.1145/1452335.1452343}
Huang, Y.; Feamster, N.; and Teixeira, R. 2008.
\newblock Practical Issues with Using Network Tomography for Fault Diagnosis.
\newblock \emph{SIGCOMM Comput. Commun. Rev.}, 38(5): 53--58.

\bibitem[{Li et~al.(2016)Li, Miao, Kim, and Yu}]{10.5555/2930611.2930632}
Li, Y.; Miao, R.; Kim, C.; and Yu, M. 2016.
\newblock FlowRadar: A Better NetFlow for Data Centers.
\newblock In \emph{Proceedings of the 13th Usenix Conference on Networked
  Systems Design and Implementation}, NSDI '16, 311--324.

\bibitem[{Liu et~al.(2019)Liu, Ben-Basat, Einziger, Kassner, Braverman,
  Friedman, and Sekar}]{10.1145/3341302.3342076}
Liu, Z.; Ben-Basat, R.; Einziger, G.; Kassner, Y.; Braverman, V.; Friedman, R.;
  and Sekar, V. 2019.
\newblock Nitrosketch: Robust and General Sketch-Based Monitoring in Software
  Switches.
\newblock In \emph{Proceedings of the ACM Special Interest Group on Data
  Communication}, SIGCOMM '19, 334--350.

\bibitem[{Liu et~al.(2016)Liu, Manousis, Vorsanger, Sekar, and
  Braverman}]{10.1145/2934872.2934906}
Liu, Z.; Manousis, A.; Vorsanger, G.; Sekar, V.; and Braverman, V. 2016.
\newblock One Sketch to Rule Them All: Rethinking Network Flow Monitoring with
  UnivMon.
\newblock In \emph{Proceedings of the 2016 ACM SIGCOMM Conference}, SIGCOMM
  '16, 101--114.

\bibitem[{McCullagh and Nelder(1989)}]{mccullagh89generalized}
McCullagh, P.; and Nelder, J.~A. 1989.
\newblock \emph{Generalized Linear Models}.
\newblock Chapman \& Hall.

\bibitem[{Moshref et~al.(2016)Moshref, Yu, Govindan, and
  Vahdat}]{10.1145/2934872.2934879}
Moshref, M.; Yu, M.; Govindan, R.; and Vahdat, A. 2016.
\newblock Trumpet: Timely and Precise Triggers in Data Centers.
\newblock In \emph{Proceedings of the 2016 ACM SIGCOMM Conference}, SIGCOMM
  '16, 129--143.

\bibitem[{{Nguyen} et~al.(2009){Nguyen}, {Teixeira}, {Thiran}, and
  {Diot}}]{5062054}
{Nguyen}, H.~X.; {Teixeira}, R.; {Thiran}, P.; and {Diot}, C. 2009.
\newblock Minimizing Probing Cost for Detecting Interface Failures: Algorithms
  and Scalability Analysis.
\newblock In \emph{IEEE INFOCOM 2009}, 1386--1394.

\bibitem[{Pazman(1986)}]{pazman86foundations}
Pazman, A. 1986.
\newblock \emph{Foundations of Optimum Experimental Design}.
\newblock Springer Netherlands.

\bibitem[{Petersen and Pedersen(2012)}]{IMM2012-03274}
Petersen, K.~B.; and Pedersen, M.~S. 2012.
\newblock The Matrix Cookbook.
\newblock Version 20121115.

\bibitem[{Pukelsheim(1993)}]{pukelsheim93optimal}
Pukelsheim, F. 1993.
\newblock \emph{Optimal Design of Experiments}.
\newblock John Wiley \& Sons.

\bibitem[{Strang(2016)}]{strang2016introduction}
Strang, G. 2016.
\newblock \emph{Introduction to Linear Algebra}.
\newblock Wellesley-Cambridge Press.

\bibitem[{Tan et~al.(2019)Tan, Jin, Guo, Zhang, Wu, Deng, Bi, and
  Xiang}]{225976}
Tan, C.; Jin, Z.; Guo, C.; Zhang, T.; Wu, H.; Deng, K.; Bi, D.; and Xiang, D.
  2019.
\newblock NetBouncer: Active Device and Link Failure Localization in Data
  Center Networks.
\newblock In \emph{16th {USENIX} Symposium on Networked Systems Design and
  Implementation ({NSDI} 19)}, 599--614.

\bibitem[{tf()}]{tensorflow}
tf. 2020.
\newblock {TensorFlow}.
\newblock https://www.tensorflow.org.

\bibitem[{{Tsang}, {Coates}, and {Nowak}(2001)}]{941208}
{Tsang}, Y.; {Coates}, M.; and {Nowak}, R. 2001.
\newblock Passive network tomography using EM algorithms.
\newblock In \emph{2001 IEEE International Conference on Acoustics, Speech, and
  Signal Processing. Proceedings (Cat. No.01CH37221)}, volume~3, 1469--1472
  vol.3.

\bibitem[{Vandenberghe and Boyd(1999)}]{vandenberghe99applications}
Vandenberghe, L.; and Boyd, S. 1999.
\newblock Applications of Semidefinite Programming.
\newblock \emph{Applied Numerical Mathematics}, 29(3): 283--299.

\bibitem[{Yang et~al.(2018)Yang, Jiang, Liu, Huang, Gong, Zhou, Miao, Li, and
  Uhlig}]{10.1145/3230543.3230544}
Yang, T.; Jiang, J.; Liu, P.; Huang, Q.; Gong, J.; Zhou, Y.; Miao, R.; Li, X.;
  and Uhlig, S. 2018.
\newblock Elastic Sketch: Adaptive and Fast Network-Wide Measurements.
\newblock In \emph{Proceedings of the 2018 Conference of the ACM Special
  Interest Group on Data Communication}, SIGCOMM '18, 561--575.

\bibitem[{{Yolanda Tsang}, {Coates}, and {Nowak}(2003)}]{1212670}
{Yolanda Tsang}; {Coates}, M.; and {Nowak}, R.~D. 2003.
\newblock Network delay tomography.
\newblock \emph{IEEE Transactions on Signal Processing}, 51(8): 2125--2136.

\bibitem[{Zhu et~al.(2015)Zhu, Kang, Cao, Greenberg, Lu, Mahajan, Maltz, Yuan,
  Zhang, Zhao, and Zheng}]{10.1145/2829988.2787483}
Zhu, Y.; Kang, N.; Cao, J.; Greenberg, A.; Lu, G.; Mahajan, R.; Maltz, D.;
  Yuan, L.; Zhang, M.; Zhao, B.~Y.; and Zheng, H. 2015.
\newblock Packet-Level Telemetry in Large Datacenter Networks.
\newblock \emph{SIGCOMM Comput. Commun. Rev.}, 45(4): 479--491.

\end{thebibliography}

\clearpage
\appendix

\section{Related Work}
\label{sec:related work}

Monitoring systems for large scale networks is a rich area. Our focus here is on systems that can scale to the size of the main cloud infrastructures (not only to one Data Center) and that can provide multiple network performance metrics with a known accuracy at all time, whatever the resources available to the telemetry systems are. We are going to address large scale telemetry systems first, and then discuss our approach and methodology.

Most recently published systems attempt to monitor the network infrastructure at full coverage (e.g.~\cite{10.1145/2934872.2934906, 10.5555/2930611.2930632, 10.1145/3230543.3230544, 10.1145/3341302.3342076, 10.1145/2829988.2787483, 10.1145/2934872.2934879, 10.1145/3230543.3230559}). These systems are generally designed for data centers and they make the assumptions that they operate with a fixed amount of resource (memory, CPU) in network nodes and/or end hosts. In practice, we claim that due to hardware diversity and operational practices (in most cloud networks, telemetry operates under a maximum budget in CPU time or probing/sampling frequency), these assumptions cannot be verified and full coverage cannot be achieved, jeopardizing the quality of the estimated metrics. This explains why we adopt a statistical approach in which network performance metrics can be estimated with a known accuracy, whatever the available resources and monitoring budget are. Compared to the approaches cited above, we do not make any assumption on minimum CPU or memory in the nodes in the network. We can operate with any probing/sampling budget, which could lead to low (but known) accuracy of our estimations.
In this landscape, NetBouncer~\cite{225976} is adopting a statistical approach for failure inference (it does not support any other metric). It relies on specific hardware to be deployed in switches (to support IP in IP tunneling) and is limited to Data Center monitoring. The approach we propose in this work uses a different optimization technique and can estimate multiple performance metrics with a know accuracy.

Our work also pertains to the area of network tomography~\cite{10.1145/511399.511338, 10.1145/2433140.2433146, 10.1145/1452335.1452343, 5461918, 5062054}. Tomography has focused historically on the theoretical (e.g.~\cite{916283, 1212670}) and practical feasibility~\cite{10.1145/1452335.1452343} of loss and latency measurements. Very often though, tomography papers deal with loss estimation and/or do not focus on optimal probing but on high quality inference. There are exceptions discussed below ; none of the papers we are aware of are dealing with the general framework approach as we define it, building a scalable implementation and evaluating it experimentally. In~\cite{941208}, the optimal sampling problem is discussed using an EM algorithm. Our statistical framework is more general as (i) it allows the estimation of network performance metrics with measurable accuracy and (ii) it only uses the network topology without requiring any traffic observations. In ~\cite{10.1145/2796314.2745862}, the authors use A- and D-optimal designs to compute probing distributions to estimate loss and delay variation. The problem statement and the methodology share similarities with our work. However, the two papers have different objectives. More importantly, ~\cite{10.1145/2796314.2745862} does not design a scalable implementation, which is the key to make these techniques applicable in practice, and as a consequence performance evaluation is limited to numerical simulations.

Our work is also related to optimal monitor placement ~\cite{4068080}. The objectives of~\cite{4068080} are quite similar to ours, as the authors try to select where and how frequently to sample traffic in network nodes in order to minimize monitoring resource consumption, and adapt to the traffic dynamics (that it takes as an input to the optimization algorithm). The method is quite different though (gradient projection for constrained optimization), and while we optimize the probing rate to estimate network performance metrics on all network links, they optimize the sampling rate in network nodes to estimate the performance of a specific set of Origin-Destination (OD) pairs. The method in ~\cite{4068080} takes the network topology and the routing matrix (i.e. which OD pair take with path) as input; it returns a set of monitors and their sampling rates that are optimal with respect to the set of OD pairs to measure. Measurement accuracy is not estimated.

A closely related work is that of \citet{DBLP:conf/sigmetrics/ChuaKC05}, where the QR decomposition of the traffic matrix is used to extract a set of \emph{important paths}. Roughly speaking, if the mean latency of these paths was known, we could estimate the latency of any other path in the network. This method is not designed for noisy measurements and we compare to it in \cref{sec:synthetic experiments,sec:real-world experiments}.

We also did not find any previous work discussing the trade-off between probing/sampling budget and performance metric measurement accuracy, and its application to monitoring very large cloud networks.

\section{Model Extensions}
\label{sec:model extensions}

\begin{figure}[t]
  \centering
  \includegraphics[width=3in]{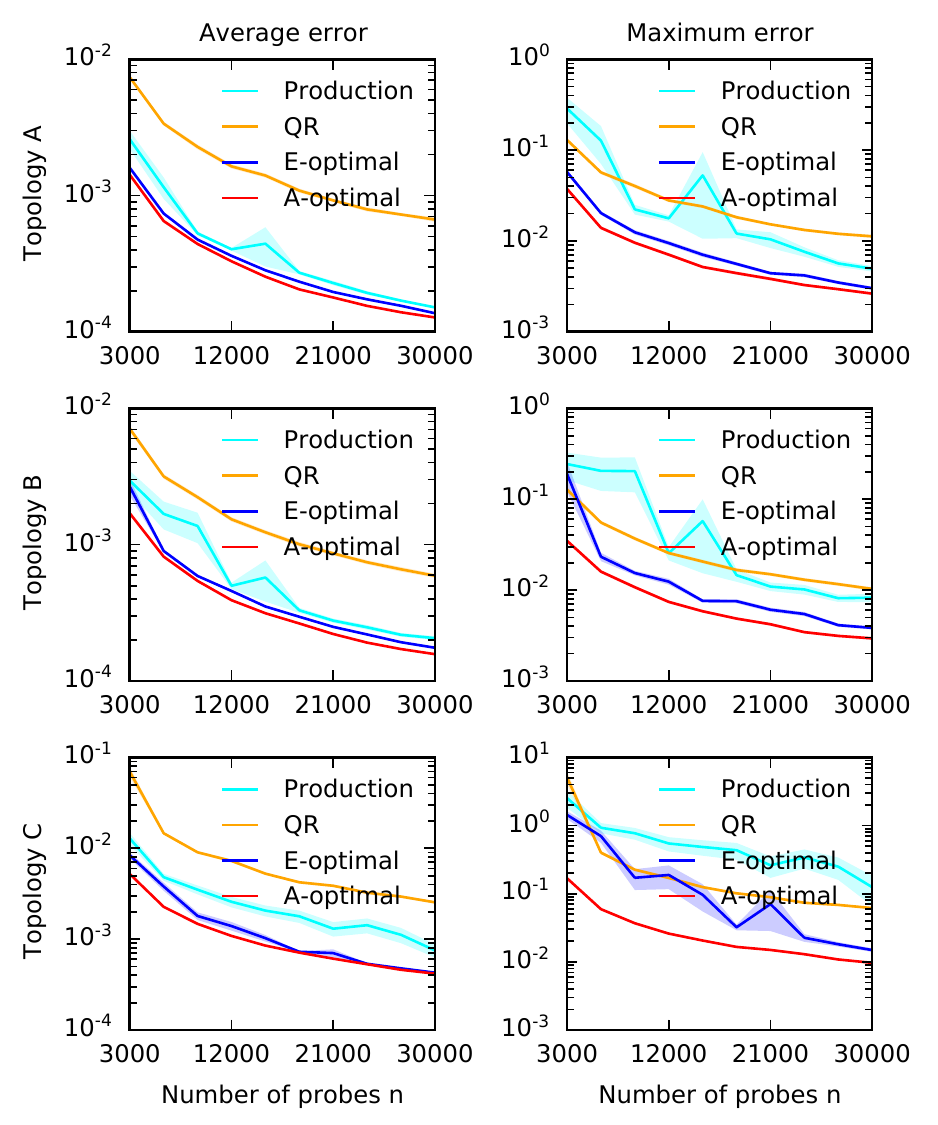}
  \caption{Packet-loss estimation errors in three network topologies. The average errors are in the first column and the maximum errors are in the second.}
  \label{fig:abc loss}
\end{figure}

\begin{figure}[t]
  \centering
  \includegraphics[width=3in]{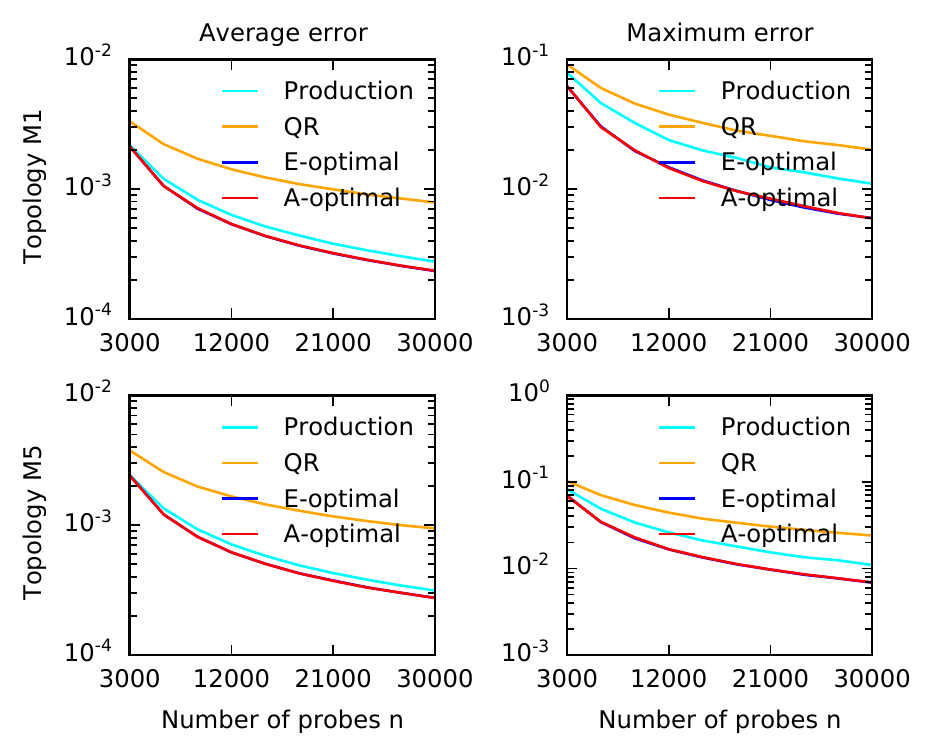}
  \caption{Packet-loss estimation errors in large-scale networks M1 and M5.}
  \label{fig:m15 loss}
\end{figure}

\subsection{Generalized Linear Models}
\label{sec:generalized linear models}

\emph{Generalized linear models (GLMs)} \citep{mccullagh89generalized} extend linear models to non-linear functions, in a way that inherits their attractive statistical and computational properties. Specifically, in the GLM, the measurement $y \in \realset$ at feature vector $x \in \cX$ has an exponential-family distribution with mean $f_\ast(x) = \mu(x\T \theta_*)$, where $\mu$ is the \emph{mean function} and $\theta_* \in \realset^d$ are model parameters \citep{mccullagh89generalized}. For any mean function $\mu$, it is well known that the matrix of second derivatives of the GLM loss at solution $\theta_*$, also called the Hessian, is
\begin{align}
  H =
  \sum_{i = 1}^n \dot{\mu}(x_i\T \theta_*) x_i x_i\T\,,
  \label{eq:hessian}
\end{align}
where $\dot{\mu}$ is the derivative of the mean function $\mu$. This Hessian plays the same role as the sample covariance matrix $G$ in \eqref{eq:squared error to covariance} \cite{chen99strong}. Hence, up to the factor of $\dot{\mu}(x_i\T \theta_*)$, the optimal designs in \cref{sec:e-optimal design,sec:a-optimal design} optimize the maximum and average errors (\cref{sec:framework}) in GLMs.

We use Poisson regression, an instance of GLMs, in \cref{sec:packet-loss experiments} to estimate packet losses. In this problem, $f_\ast(x) = \exp[x\T \theta_\ast]$ is the probability that the packet is not dropped on path $x \in \cX$ and $\theta_\ast \in \realset^d$ is a vector of log-probabilities that packets are not dropped on individual path edges.

\subsection{Non-Linear Models}
\label{sec:non-linear models}

To extend our ideas to arbitrary non-linear models, we propose learning a non-linear mapping of features into a lower $r$-dimensional space, so that we get a linear regression problem in the new space. That is, we learn $g: \realset^d \to \realset^r$ such that $f_\ast(x) \approx g(x)\T \theta_\ast$.

The function $g$ is learned as follows. Let $m$ be the number of tasks and $\cD_j$ be the dataset corresponding to task $j \in [m]$. In networking, the task $j$ can be viewed as an inference problem on day $j$, which is accompanied by its training set $\cD_j$. We learn $g$ by solving
\begin{align}
  \min_g \min_{\theta_1, \dots, \theta_m}
  \sum_{j = 1}^m \sum_{(x, y) \in \cD_j}
  (g(x)\T \theta_j - y)^2\,,
  \label{eq:embedding learning}
\end{align}
where $\theta_j \in \realset^r$ are optimized model parameter for task $j$.

The key structure in above loss function is that function $g$ is shared among the tasks. Therefore, its minimization leads to learning a common embedding $g$, essentially a compression of features $x$, that is useful for solving all tasks using linear regression. The above optimization problem can be solved using a multi-headed neural network, where $g$ is the body of the network and head $j$ outputs $g(x)\T \theta_j$, the predicted value at feature vector $x$ for task $j$.

\subsection{Local Budget Constraints}
\label{sec:local budget constraints}

Both optimal designs \eqref{eq:e-optimal design} and \eqref{eq:a-optimal design} output a probing distribution $\alpha = (\alpha_x)_{x \in \cX}$ subject to a single global constraint $\sum_{x \in \cX} \alpha_x = 1$. In practice, local constraints are common. For instance, we may want to enforce that most paths do not start in a single source node, simply because such a probing strategy cannot be implemented due to resource availability constraints. Such constraints can be enforced as follows. Let $\mathrm{src}(x)$ be the source node of path $x$ and $S = \set{\mathrm{src}(x): x \in \cX}$ be the set of all sources. Then
\begin{align*}
  \forall s \in S:
  \sum_{x \in \cX} \I{\mathrm{src}(x) = s} \alpha_x \leq b
\end{align*}
is a set of linear constraints in $\alpha$ that limit the probing probability from any source $s$ to at most $b$. Because the constraints are linear in $\alpha$, they can be easily incorporated into \eqref{eq:e-optimal design} and \eqref{eq:a-optimal design} without changing the hardness of these problems. We experiment with local budget constraints in \cref{sec:local budget experiments}.

\section{Additional Experiments}
\label{sec:additional experiments}

\begin{figure*}[t]
  \centering
  \includegraphics[width=6in]{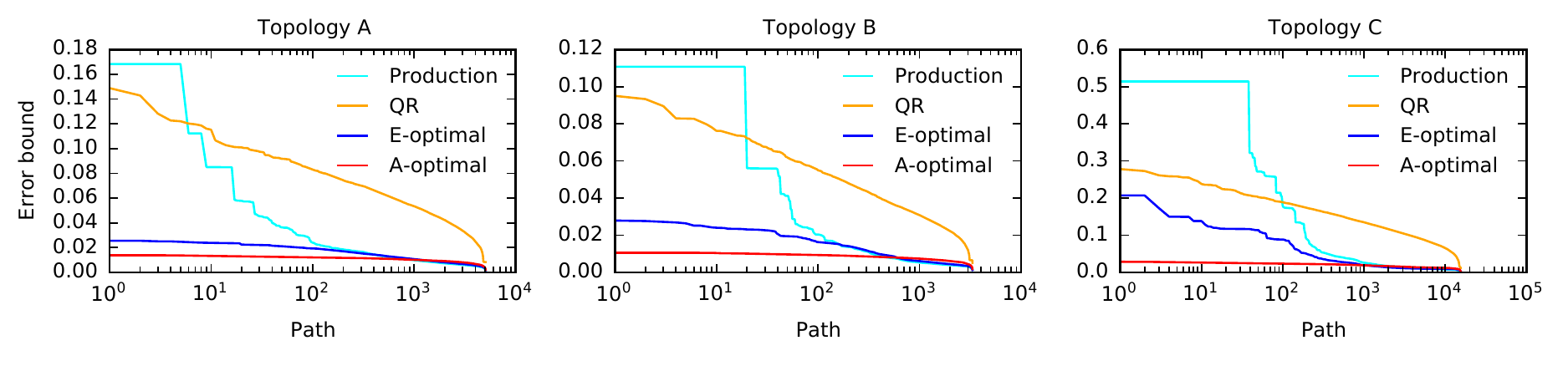}
  \vspace{-0.1in}
  \caption{Predicted errors in \eqref{eq:squared error to covariance}, in the descending order of the per-path errors for each method. The $x$-axis is on the logarithmic scale to highlight highest predicted errors, which contribute the most to the actual errors.}
  \label{fig:abc allocation}
\end{figure*}

\begin{figure}[t]
  \centering
  \includegraphics[width=3in]{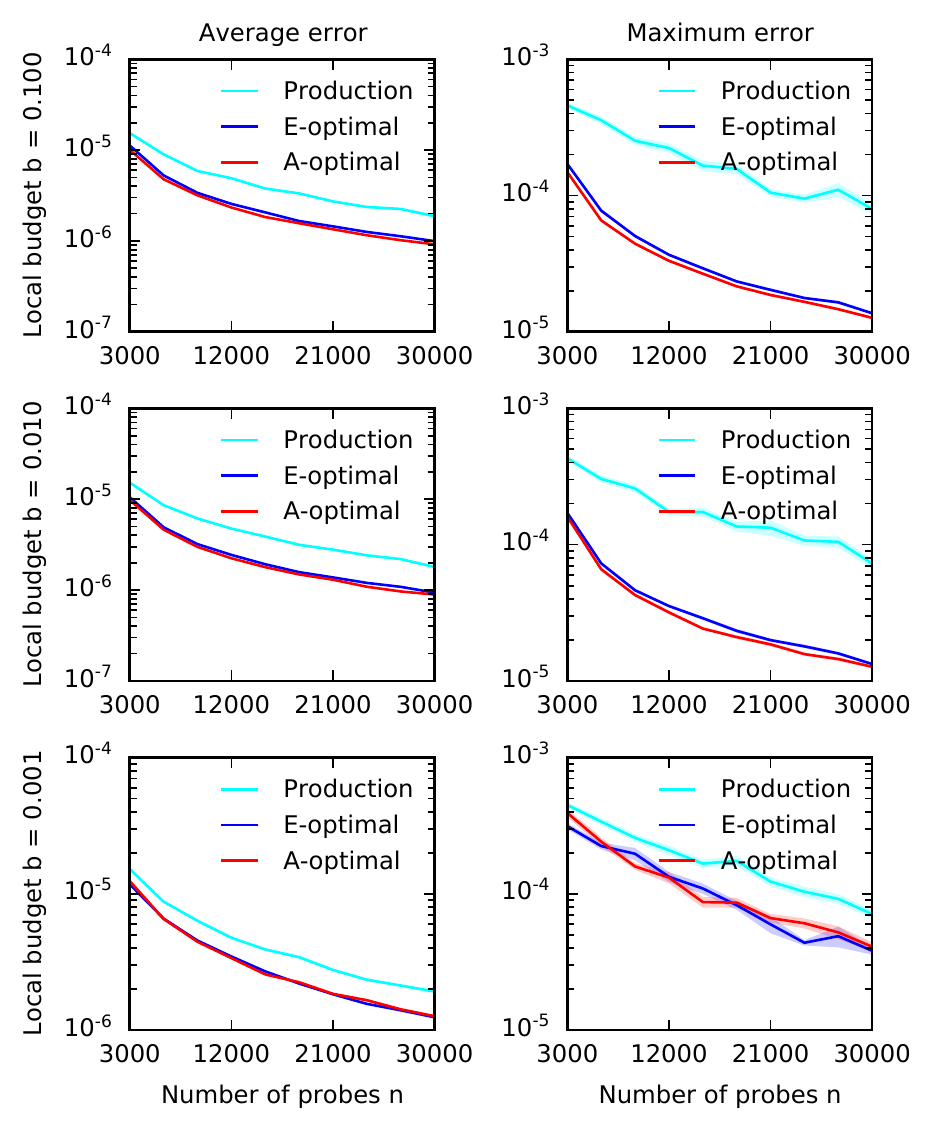}
  \caption{Latency estimation errors in topology A for various local budgets. The average errors are in the first column and the maximum errors are in the second.}
  \label{fig:local budget latency}
\end{figure}

\begin{figure*}[t]
  \centering
  \includegraphics[width=6in]{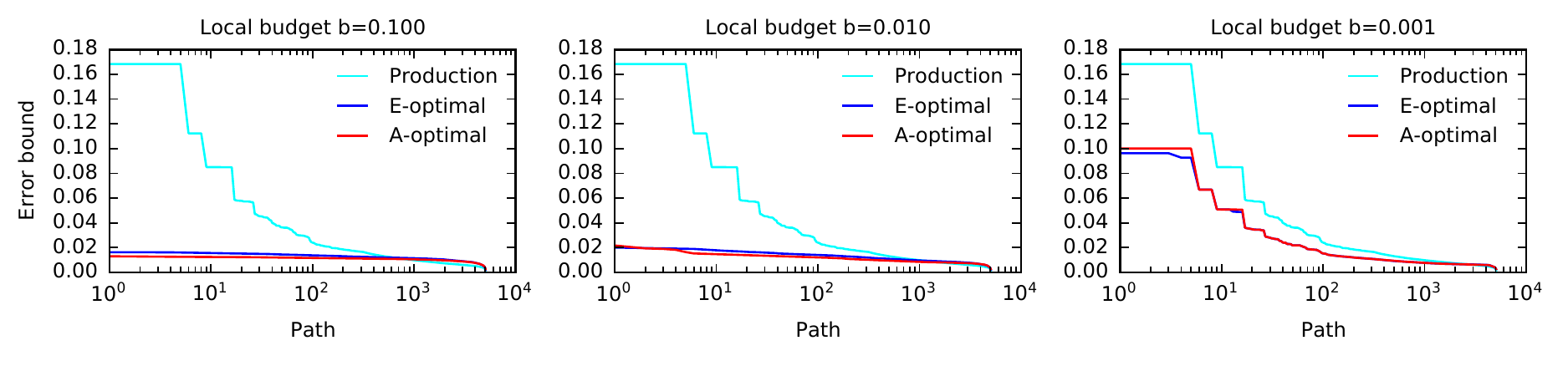}
  \vspace{-0.1in}
  \caption{Predicted errors in \eqref{eq:squared error to covariance}, in the descending order of the per-path errors for each method. The $x$-axis is on the logarithmic scale to highlight highest predicted errors, which contribute the most to the actual errors. The plots are for topology A and various local budgets.}
  \label{fig:local budget allocation}
\end{figure*}

\subsection{Synthetic Packet-Loss Experiments}
\label{sec:packet-loss experiments}

\noindent \textbf{Packet Loss Simulator}. We denote by $\theta_*(e)$ the log-probability that the packet is not dropped on edge $e$ and by $\theta_* \in \realset^d$ all edge log-probabilities. So $\exp[\theta_*(e)]$ is the probability that the packet is not dropped on edge $e$. Using the mean latency of edge $e$ in \cref{sec:latency experiments}, $\ell(e)$, we define $\theta_*(e)$ as
\begin{align*}
  \theta_*(e)
  = - \frac{\ell(e)}{10 \max_{e \in [d]} \ell(e)}\,.
\end{align*}
This means that longer edges are more likely to lose packets. This is simply a modeling choice to generate edges with variable losses. By definition, $\theta_*(e) \in [-0.1, 0]$ and thus $\exp[\theta_*(e)] \in [0.9, 1]$. So the probability of dropping a packet on any edge is at most $0.1$.

As in \cref{sec:latency experiments}, $x$ is a vector of edge indicators in a path. Thus $f_*(x) = \exp[x\T \theta_*]$ is the probability that the packet is not dropped on path $x$, under the assumption that the events of dropping packets on path edges are independent. The \emph{noisy packet-loss observation} of path $x$ is $y \sim \mathrm{Ber}(f_*(x))$, an indicator that the packet is not dropped with mean $f_*(x)$. Note that estimation of $\theta_*$ can be solved by Poisson regression, an instance of GLMs (\cref{sec:generalized linear models}). The rest of the experimental setup is the same as in \cref{sec:latency experiments}.

\noindent \textbf{Results}. Packet-loss estimation errors are reported in \cref{fig:abc loss}. These results are consistent with those described earlier for latency (see \cref{fig:abc latency}). In particular, we observe that the errors decrease as budget $n$ increases and that the rate of this decrease is inversely proportional to the budget. That is, from budget $n = 3\,000$ to $n = 30\,000$, the decrease is also about $10$ fold. Such a good result is surprising, since \eopt and \aopt minimize the average and maximum errors in GLMs only loosely, as discussed in \cref{sec:generalized linear models}. We also observe that the reported errors are acceptable. For instance, in all network topologies, the average squared error of \aopt with budget $n = 30\,000$ is below $5 \times 10^{-4}$. This means that the absolute path loss is mispredicted by $2 \times 10^{-2}$ on average, which is nearly an order of magnitude lower than the maximum packet loss on an edge, which is $0.1$ by our modeling assumptions.

We also observe that \eopt and \aopt consistently dominate both \markbaseline and \uniform. In particular, they have lower errors than these baselines for all budgets and attain any fixed error with a lower budget than either of them. We observe this trend for both the average and maximum errors, and over all network topologies. The differences between \eopt and \aopt are less pronounced than in \cref{fig:abc latency}, although clearly \aopt never performs much worse than \eopt.

\subsection{Real-World Packet-Loss Experiments}
\label{sec:real-world packet-loss experiments}

For the large-scale topologies used in the real-world experiments (described earlier for latency in \cref{sec:real-world experiments}), we observe packet loss estimation results similar to those described for latency in \cref{sec:real-world results}. Refer to \cref{fig:m15 loss} for the packet loss experiment results.

These results are consistent with the trends observed in the synthetic experiments describe abpve. Specifically, the errors decrease as budget $n$ increases, and \eopt and \aopt consistently dominate both the baselines (\markbaseline and \uniform). For packet loss, the average squared error of \aopt with budget $n = 30\,000$ is below $2 \times 10^{-4}$, which means that the absolute path loss is mispredicted by $10^{-2}$ on average. Both absolute errors are one order of magnitude lower than the estimated quantities, which means that our predictions are reasonably accurate.

\subsection{Prediction Error Explained}
\label{sec:prediction error explained}

Our next step is to better understand why \aopt performs so well and why \markbaseline performs so poorly. To do so, we use the error bound in \eqref{eq:squared error to covariance}. For a given method and path, the error bound is the maximum high-probability error for latency prediction of that path by that method. A lower predicted error should correlate with a higher accuracy of latency estimates. Therefore, if one method is more accurate than another, we would expect most of its predicted errors to be lower. To visualize this, we show the predicted errors of all paths for each method in \cref{fig:abc allocation}, in the descending order of the per-path errors. We observe that the \aopt predicted errors are lower overall than those of \eopt, which are lower overall than those of \uniform. The predicted errors of \markbaseline are the highest, which indicates poor performance. This is consistent with the actual errors in \cref{fig:abc latency}. We conclude that the optimization of \eqref{eq:squared error to covariance}, which is done in both \eopt and \aopt, leads to better performance.

\subsection{Local Budget Experiments}
\label{sec:local budget experiments}

The goal of this section is to evaluate local budget constraints described in \cref{sec:local budget constraints}. We consider the most common constraint in network monitoring, that none of the nodes in the network is a source or destination of \say{too many} probes. This constraint can be incorporated in our optimal designs as follows. Let $\cV$ be the set of all nodes in the network and $\mathrm{src}(x)$ be the source of path $x$. Then, for each $v \in \cV$, the constraint is
\begin{align*}
  \sum_{x \in \cX} \I{\mathrm{src}(x) = v} \alpha_x
  \leq \frac{\sum_{x \in \cX} \I{\mathrm{src}(x) = v}}{\abs{\cX}} + b\,,
\end{align*}
where $b \geq 0$ is an \emph{excess local budget} over uniform probing distribution and $\sum_{x \in \cX} \I{\mathrm{src}(x) = v} / \abs{\cX}$ is the fraction of paths starting in node $v$. The destination constraints are defined analogously.

We experiment on topology A with three local budgets $b \in \set{0.1, 0.01, 0.001}$. The latency estimation results are reported in \cref{fig:local budget latency}. We expect that lower local budgets would result in worse optimal designs, because both \eopt and \aopt are more constrained in how they can probe. We observe this with the maximum error at $n = 30\,000$, which increases from $10^{-5}$ ($b = 0.1$) to $3 \times 10^{-5}$ ($b = 0.001$). We also observe slight changes in the average error, which increases from slightly below $10^{-6}$ ($b = 0.1$) to slightly above $10^{-6}$ ($b = 0.001$).

In \cref{fig:local budget allocation}, we show the predicted errors for latency and various local budgets. This plot is similar to \cref{fig:abc allocation}. When the local budget is low ($b = 0.001$), the probing distributions in \eopt and \aopt cannot differ much from \uniform, and some paths have high predicted errors of $0.1$. This leads to lower accuracy. On the other hand, when the local budget is high ($b = 0.1$), the probes can be distributed more intelligently and this results in the highest predicted errors of mere $0.02$, similarly to no local budget constraints (topology A in \cref{fig:abc allocation}).

We conclude that our approach can easily incorporate local budget constraints. Perhaps surprisingly, even when the local budget is $b = 0.01$ (which means that no more than $1\%$ of the probed paths can start or end at any given node) the corresponding \eopt and \aopt solutions offer similar performance as with global budget only. 

\end{document}